\documentclass[10pt,twocolumn,letterpaper]{article}

\usepackage{cvpr}
\usepackage{times}
\usepackage{epsfig}
\usepackage{graphicx}
\usepackage{amsmath}
\usepackage{amssymb}

\usepackage[pagebackref=true,breaklinks=true,letterpaper=true,colorlinks,bookmarks=false]{hyperref}

\cvprfinalcopy % *** Uncomment this line for the final submission

\def\cvprPaperID{****} % *** Enter the CVPR Paper ID here

\ifcvprfinal\fi
\begin{document}

%%%%%%%%% TITLE
\title{Insta(nt) Pet Therapy: GAN-generated Images for Therapeutic Social Media Content}

\author{Tanish Jain\\
Stanford University\\
{\tt\small tanishj@stanford.edu}
% \and
}

\maketitle
%\thispagestyle{empty}

%%%%%%%%% ABSTRACT
\begin{abstract}

% Concise summary of the problem, approach, and key results. Less than 300 words.

The positive therapeutic effect of viewing pet images online has been well-studied. However, it is difficult to obtain large-scale production of such content since it relies on pet owners to capture photographs and upload them. I use a Generative Adversarial Network-based framework for the creation of fake pet images at scale. These images are uploaded on an Instagram account where they drive user engagement at levels comparable to those seen with images from accounts with traditional pet photographs, underlining the applicability of the framework to be used for pet-therapy social media content.
\end{abstract}

%%%%%%%%% BODY TEXT
\section{Introduction}

% Introduction (2 points): Outline the problem your team worked on, its significance, and provide an overview of your results.

Several studies have highlighted the positive impact of viewing images of pets on social media apps such as Instagram \cite{riddle2020social} \cite{zhou2020sense}. Zhou et al., for instance, demonstrate ``the positive effect of an online pet on subjective well-being'' \cite{zhou2020sense}, underlining its therapeutic effect. However, content may not be necessarily tailored for this purpose, and users can only see these images as frequently as the pet owner chooses to post them.

In this paper, I build on prior work to explore two different GAN architectures and their ability to generate pet images. Importantly, the images must be of a quality that can drive user engagement on social media, which can indicate the presence of the desired therapeutic effect. 

I add to the existing literature by using a new dataset which results in better model performance. Additionally, I use the generated images for a novel task by using them directly on an Instagram account and evaluate their ability to create engagement. The results are promising: GAN-generated images are able to reach comparable or better levels of engagement when compared to real images. This showcases the applicability of the approach for the generation of therapeutic social media content in the context of online pet watching.

\section{Related Work}\label{sec:relatedwork}

% Background/Related Work (8 points): Cover all of background, related works (minimum of 5), and pitfalls (if applicable) in detail.

Prior work has shown that therapy dog pages on Instagram perform well in terms of generating user engagement, which is backed by the positive effects of ``online pet watching'' \cite{zhou2020sense}. Blaine and Kremer \cite{blaine2018we}, for example, discuss this phenomenon; however, the difficulty of generating this in-demand content is highlighted, especially since it relies on pet owners consistently uploading photographs of their pets on an online platform. In this project, I attempt to solve this problem with the use of GAN-generated images, which may be produced on a large scale effectively.

Although the application itself is novel, several prior works have looked at the problem of using GANs for generating images of dogs. Shangguan et al. \cite{shangguan2020dog} utilize a DCGAN to generate images of dogs. The authors utilize the Stanford Dogs dataset, which I replace in my approach for reasons discussed in Section \ref{sec:dataset}. Kumari et al. \cite{kumari2021dcgan} also use a DCGAN to generate dog images. Their model is pretrained on ImageNet, which is shown to work well and produce a more diverse set of images (namely, by producing dogs of various breeds). Both these approaches, however, generate low-resolution images, which would be unfit for an Instagram-based solution. 

Compared to the problem of generating dog images, the problem of generating cat images has been relatively less explored. While the high-level problem is not fundamentally different, the performance of models in prior work across these two tasks is not consistent, which may underline a problem with the choice of the training dataset. Gupta and Gupta \cite{gupta2021performance} use two models - DCGAN and LS-GAN - to generate cat images. They note that the DCGAN produces realistic images, which is critical for the proposed application. 

Prior work on the generation of animal images suggests a strong preference for the use of DCGAN; however, concerns around its ability to produce high-resolution images remain. Jain et al. \cite{jain2020performance} have shown that an alternative model, BigGAN, can produce better quality results than DCGAN. However, they note that the model training can be much more time-consuming and resource-intensive, which may explain the preference for DCGAN.

In my project, I attempt to build on this prior work and combine individual improvements to comprehensively address some of the underlined challenges. I use a new dataset to address issues arising from the choice of dataset. This is further discussed in Section \ref{sec:dataset}. I use both a DCGAN and a BigGAN given their individual advantages and compare their results. I augment data by using images from two datasets, based on its benefits noted in \cite{shangguan2020dog}, and fine-tune a pretrained model in the case of BigGAN, instead of training from scratch, to address concerns around computational constraints.

\section{Dataset}\label{sec:dataset}

% Dataset (4 points): Outline the type of data you used, where/how you obtained it, the size of the dataset, the train/validation/test splits, and preprocessing (if any) done on the dataset. Include basic stats, visualizations, and limitations if any.

The project uses the ``Dogs vs Cats'' Kaggle dataset\footnote{https://www.kaggle.com/c/dogs-vs-cats/data} as well as the ``Cats and Dogs Breeds Classification Oxford Dataset''\footnote{https://www.kaggle.com/zippyz/cats-and-dogs-breeds-classification-oxford-dataset} for this task. The former contains 25,000 images in total, of which exactly half are images of dogs, and half are images of cats. The latter contains 7,393 images in total, where the ratio of cat images to dog images is approximately 2:1. 

These datasets have not been used for pet image generation in existing literature, which therefore presents an opportunity to evaluate the impact of the choice of the dataset on the quality of generated images. Other popular datasets, and especially the Stanford Dogs dataset, which has been a popular choice for using with GANs before, poses some challenges. This includes the fact that the images in the dataset often contain wider shots which may contain less interesting features, and the dogs themselves may form a small part of the image. The new datasets also contains images of both dogs and cats, the two most popular pet types on the internet \cite{riddle2020social}.

% Add Image: Preprocessing
% Add Image: Image samples from datasets
\begin{figure}[t]
  \centering
   \includegraphics[width=0.99\linewidth]{ 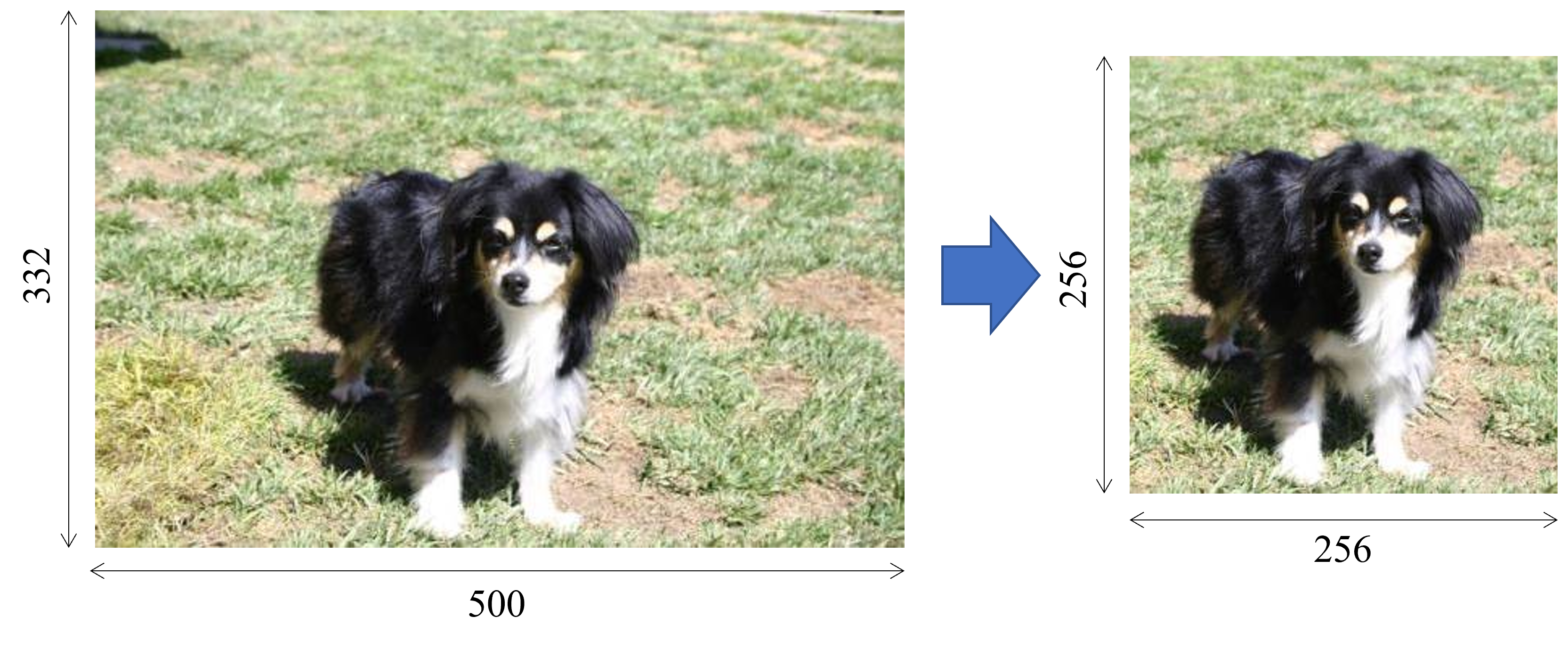}

   \caption{Acceptable images are cropped/resized during preprocessing.}
   \label{fig:preprocessing}
\end{figure}

Nevertheless, there are a few limitations of the chosen datasets. In the ``Cats vs. Dogs'' dataset, although the number of cat and dog images is the same, and diversity of the cat images is lower than that of the dog images in terms of the number of unique breeds and frequency of images for each breed, even when accounting for the fact that dog breeds are more numerous in general. Additionally, images may be of different sizes and may contain multiple animals and humans. However, these images are annotated, making it possible to crop and filter images using data preprocessing. The ``Cats and Dogs Breeds Classification Oxford Dataset'' contains much higher quality images overall in terms of resolution and focus on the animal; however, it contains fewer images. The data from both these datasets is combined to ensure data diversity while ensuring that enough images of acceptable quality are retained after preprocessing. 

% Add Image: Image samples from datasets
\begin{figure}[t]
  \centering
   \includegraphics[width=0.99\linewidth]{ 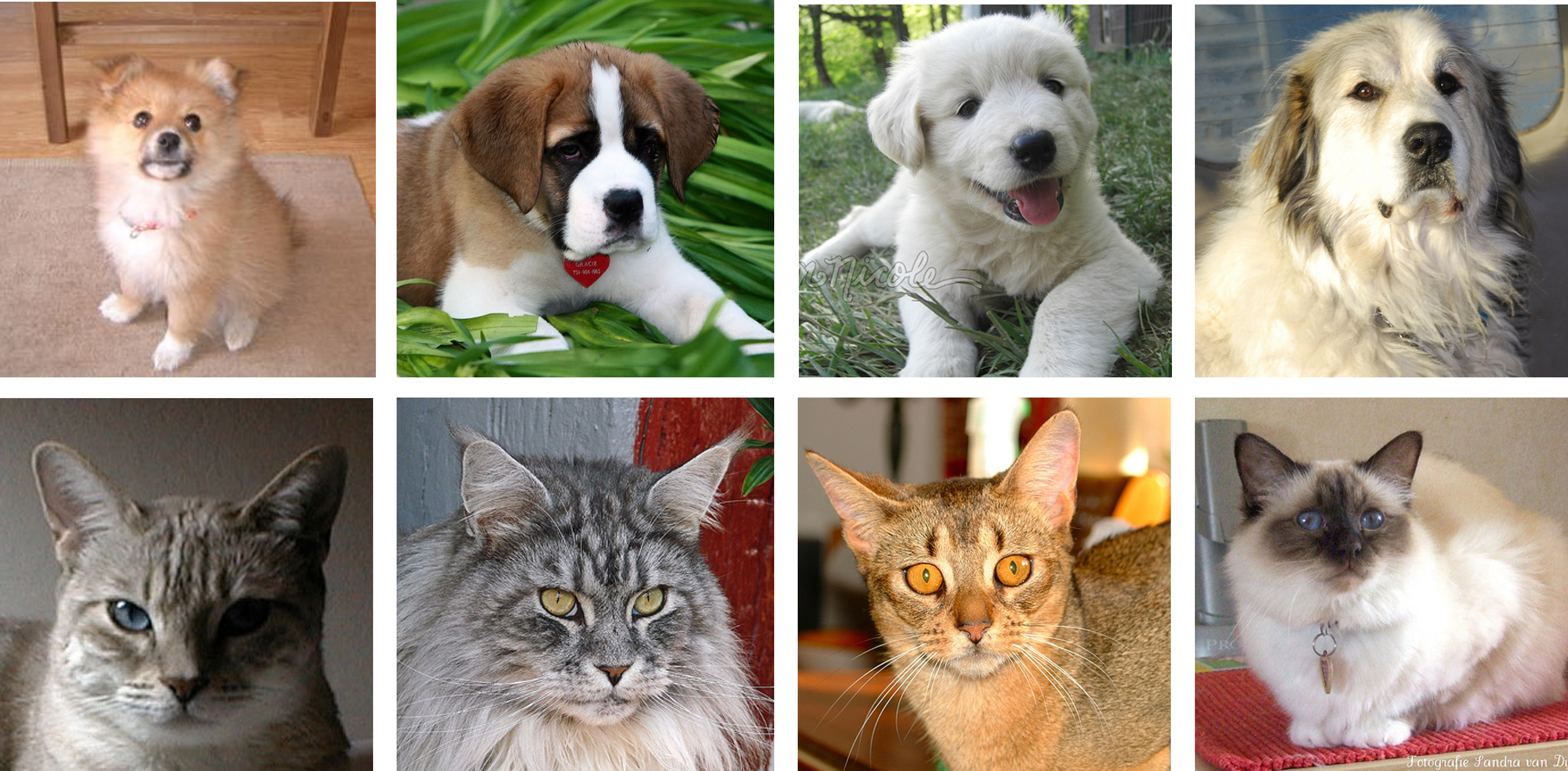}

   \caption{Sample images from curated dataset after preprocessing.}
   \label{fig:sample}
\end{figure}

For the preprocessing step, images with low-resolution (less than $256$ px in either dimension) and humans were dropped entirely. Then, using image annotations, each image was cropped into a square shape focused on the pet and resized to either $64\times 64$ or $256\times 256$ pixels, depending on the model. The data was augmented by adding laterally-inverted copies of each image. Finally, the images were normalized. In total, 36,232 images were available for use after preprocessing, with dog and cat images represented equally.

\section{Methods}\label{sec:methods}

% Methods (6 points): Discuss and describe your approach and justify why it was used over alternate methods. Methods and approaches should be explained clearly supplemented with equations and diagrams.

I utilize two existing GAN architectures - DCGAN and BigGAN - to develop a model for the specific task of generating high-quality pet images for Instagram. These models are trained using the curated dataset discussed in Section \ref{sec:dataset}. Additionally, the user engagement generated by these images is evaluated.

% DCGAN
\subsection{DCGAN}

For the baseline model, I use a Deep Convolutional Generative Adversarial Network (DCGAN) architecture. DCGAN contains a Generator and a Discriminator like a traditional GAN, but both Generator and Discriminator models are based on convolutional neural networks. \cite{radford2015unsupervised}. This was chosen as the baseline as it has been widely used in prior work \cite{shangguan2020dog}\cite{kumari2021dcgan} and may be trained using moderate computational resources \cite{jain2020performance}.

The implementation is largely based on the model and techniques described by Radford et. al. in their paper introducing DCGAN \cite{radford2015unsupervised}. The model was implemented using \cite{radford2015unsupervised} and the DCGAN implementation in Coursera's ``Build Basic Generative Adversarial Networks'' course \cite{zhou_zhou_zelikman}. 

%% Add Image: DCGAN

To train this model, I use a version of the binary crossentropy loss, defined by:

\begin{equation*}
\resizebox{0.45\textwidth}{!}{$ V(D,G) = {E_{x\sim {P_{data}}(x)}}[\log D(x)] + {E_{z\sim {P_z}(z)}}[\log (1 - D(G(z)))] $}
\end{equation*}

Where D and G represent Discriminator and Generator. $G(z)$ is a sample generated by a random vector and x is real sample data. The optimal weight value is obtained by minimizing the loss function.

% BigGAN
\subsection{BigGAN}

\begin{figure}[t]
  \centering
   \includegraphics[width=0.99\linewidth]{ 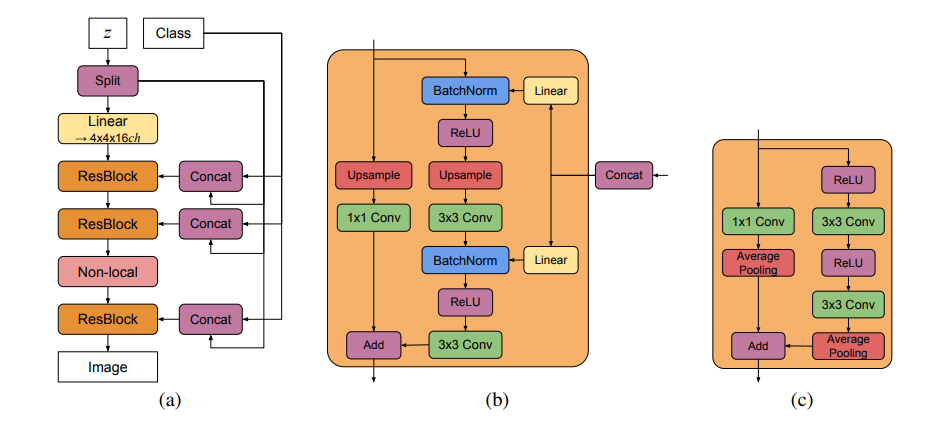}

   \caption{(a) A typical architectural layout for BigGAN’s G.
(b) A Residual Block (ResBlock up) in BigGAN’s G. (c) A Residual Block (ResBlock down) in
BigGAN’s D. Source: Brock et al. \cite{brock2018large}}
   \label{fig:biggan}
\end{figure}

BigGAN is a GAN architecture based on Self-Attention GAN that is capable of synthesizing high-fidelity images, shown in Figure \ref{fig:biggan} \cite{brock2018large}. An important component in the model is the use of the "truncation trick" which when used during image generation results in an improvement in image quality \cite{jain2020performance}. This model also uses orthogonal regularization:

\begin{equation*}
{R_\beta }(W) = \beta \left\| {{W^ \top }W \odot (1 - I)} \right\|_{\text{F}}^2\end
{equation*}

Where $\beta$ is a hyperparameter and $W$ is weight matrix, which improves the possibility of the truncation trick in generating high quality images \cite{jain2020performance}. However, compared to DCGAN, training the model is computationally expensive and time-consuming.

%% Add Image: BigGAN

I rely on the official implementation provided by the authors, which has been pretrained on ImageNet.\footnote{https://github.com/ajbrock/BigGAN-PyTorch} The use of a pretrained model can result in significant time and resource savings, making it ideal given the resource constraints. The pretrained model is then fine-tuned with my curated dataset.

\subsection{Evaluation}

\subsubsection{Model Evaluation}

I use the Inception Score (IS) for evaluating model performance. The IS incorporates measures of both fidelity and diversity \cite{salimans2016improved}, which is relevant for my task. 

Additionally, since prior work on related tasks uses IS, it is possible to compare results for my models.

\subsubsection{User Engagement Evaluation}
% Instagram engagement
To evaluate the applicability of the approach for the specific task of generating images for social media use, it is critical to measure user engagement generated by the images using a metric. Importantly, this metric must be simple, application-specific (in this case, suitable for Instagram) and use publicly-available data to allow for comparison with content from other pages on the platform. 

Yew et al. \cite{yew-insta} propose the ``Instagram Engagement Score'' or IES using marketing case studies, which is a simple yet effective metric for measuring engagement. I use a scaled version of this metric, given by the formula:

\begin{equation*}
\begin{array}{c}Instagram\,Engagement\,Score \\ = \displaystyle\frac{(\text{Number} \, \text{of} \, \text{Likes} + \text{Number} \, \text{of} \, \text{Comments})}{\text{Number} \, \text{of} \, \text{Followers}} \end{array}
\end{equation*}

Critically, this metric can capture engagement across pages of different follower counts and uses publicly available data (i.e., the number of likes, comments, and followers). The metric may be applied to measure both image-level and page-level engagement. In this study, the image-level IES (or i-IES) is measured exactly 24 hours after an image is posted, using the follower count of the page and likes and comments count of the image of interest. The page-level IES (or p-IES) is measured at any given instant, using the follower count of the page and the sum of likes and comments count of the 10 most recent relevant images on the page.

%%%%%%%%%%%

\section{Experiments}

% Experiments (8 points): Outline the experiments which were performed to evaluate your approach. This could include an ablation study to demonstrate the impact of each of your methods (this is what your baseline is for), a hyper parameter search such as lambda values of loss function components, etc. If you tried to improve an existing model architecture, you could compare your experimental results with those of published works. Include graphs, tables, and figures to show your experimental results. Include the description and results of your risky experiments here! If it went well, congratulations! If it didn’t, that’s okay! Try to rationalize why your experiment may not have succeeded.

\subsection{Image Generation}

Images were generated using both the DCGAN and BigGAN models discussed in Section \ref{sec:methods}. For both models, data was preprocessed and augmented using the approach outlines in Section \ref{sec:dataset}. Although the models were originally trained with dog images only, they were later trained with both dog and cat images in order to expand the scope of the project.

As the baseline, a DCGAN model was trained with the dataset from scratch. Following the recommendations in \cite{radford2015unsupervised}, I used the Adam optimizer for both the generator and the discriminator, with learning rate, $lr = 0.0002$ and $\beta_1 = 0.5$. The DCGAN was trained with smaller-sized $64\times 64$ images using a single GPU setup on the Farmshare 2 environment of the Stanford Research Computing Center. The model took a long time to train, and the fidelity of images, unsurprisingly, improved with the number of epochs.

%% ADD IMAGES: DCGAN Examples

The generated images had an IS score of 10.399 (dog images only) and 7.484 (dog and cat images) after 500 epochs. Notably, this is higher than the IS score achieved by models in prior work \cite{jain2020performance}, potentially underlining the impact of the dataset choice.

Nevertheless, the images generated were low-resolution and easily distinguishable as artificial by humans, making them unfit for use as social media content. A BigGAN model was used next given its ability to produce high-quality images \cite{brock2018large}. While BigGAN can produce better-quality images, it also requires significantly more time and resources to train \cite{jain2020performance}. As a result, I used an implementation pretrained on ImageNet provided by Brock et al. \cite{brock2018large}. This is then fine-tuned with my application-specific dataset using the larger $256\times 256$ images.

Following the recommendations in \cite{brock2018large} and given the resource constraints, I use the Adam optimizer for both the generator and the discriminator, with learning rates, $lr = 0.0001$ and $0.0004$, respectively with a batch size of 256. The model was trained (fine-tuned) for 10,000 iterations using a 4-GPU setup on Farmshare 2.

%% ADD IMAGES: BigGAN Examples

The model generated images that were not only higher-resolution but also more realistic than the baseline approach. Additionally, the model had an IS of 94.331 (dog images only) and 67.500 (dog and cat images), which presents a significant improvement over the baseline as well as the computationally-efficient approaches in prior works discussed in Section \ref{sec:relatedwork}.

\subsection{Managing an Instagram Page}

\begin{figure}[t]
  \centering
   \includegraphics[width=0.99\linewidth]{ 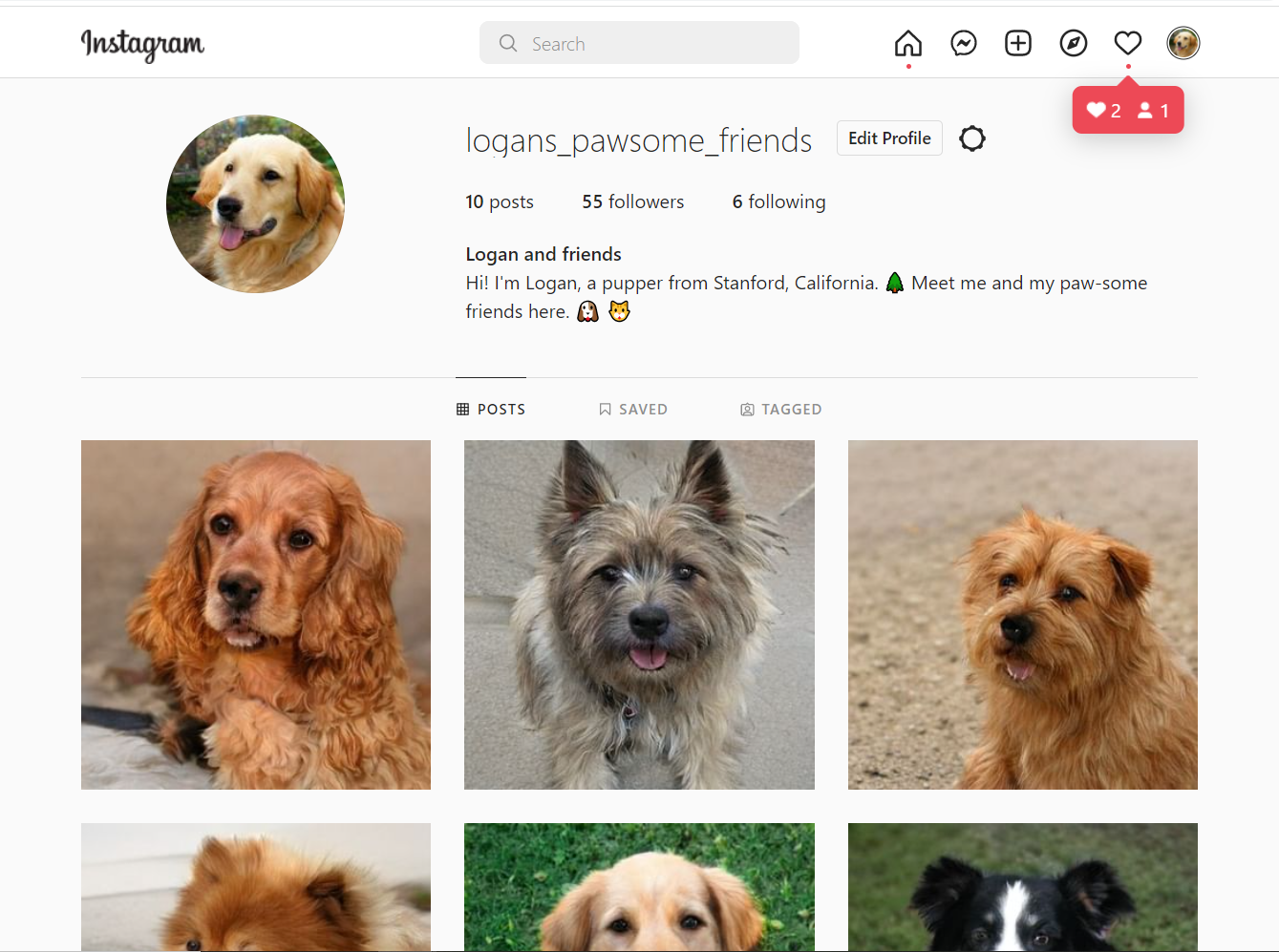}
   \caption{Screenshot of @logans\_pawsome\_friends Instagram page.}
   \label{fig:screenshot}
\end{figure}

An Instagram page was set up to upload generated images, using the handle @logans\_pawsome\_friends. The page follows the life of the titular golden retriever, Logan, who meets his various animal friends and posts their images. Since the goal is to evaluate engagement produced by GAN-generated content, no indication is provided about the images being synthetic, outside of subtle hints, such as the name of characters (such as LoGAN) and cryptic captions (``A fun day out with Snowy - it was \textit{unreal}!''). A screenshot of the page is shown in Figure \ref{fig:screenshot}.

%% ADD Image: Insta screenshot

\begin{figure}[t]
  \centering
   \includegraphics[width=0.99\linewidth]{ 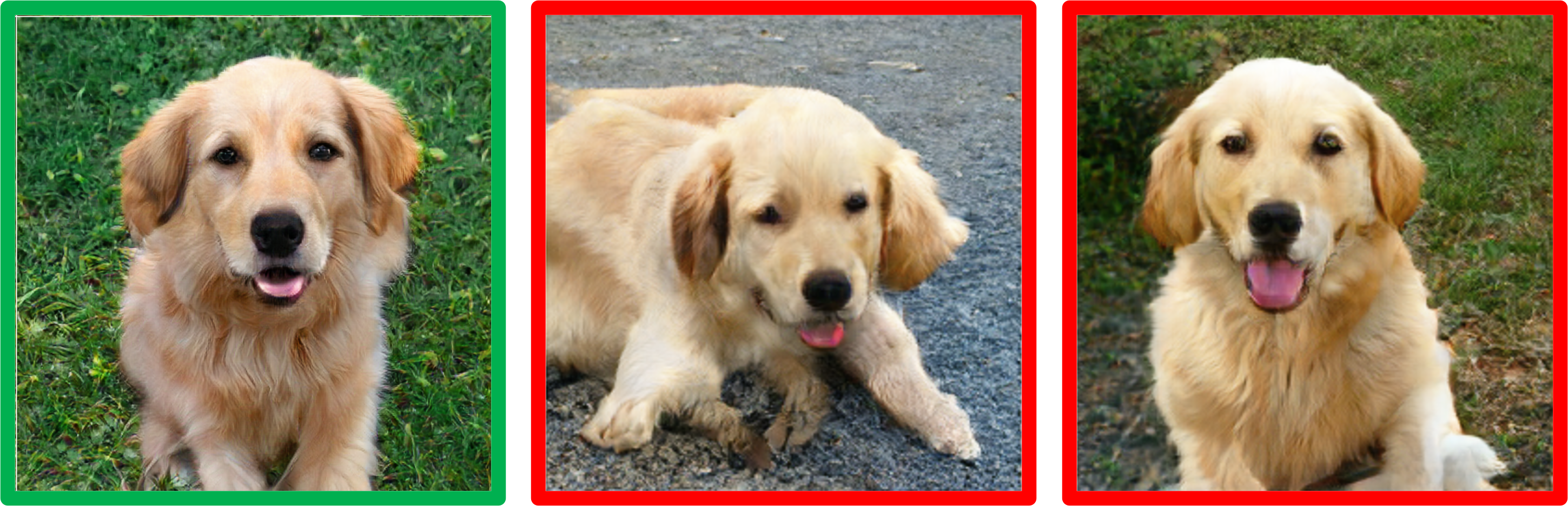}

   \caption{The top images were selected manually from the set of GAN-generated images for posting on the Instagram page; above, only the left-most image was considered fit for uploading.}
   \label{fig:selection}
\end{figure}

While images generated by the model are highly realistic, they may still be distinguishable by humans. As a result, images for posting are manually picked. As of the writing of this report, 15.2\% of all generated dog samples have been considered fit for posting on the page. An example of the selection process is shown in Figure \ref{fig:selection}. Surprisingly, none of the cat images generated were fit for posting; while many images are high-quality, they are still distinguishable by humans.

%% ADD Image: Selection process
%% ADD IMAGE: Images with highest IES

For evaluating the engagement of the images of @logans\_pawsome\_friends with other Instagram pages, three categories of pages are delineated: ``low popularity'' pages with under 100 followers, ``medium popularity pages with between 100 and 10,000 followers, and ``high popularity'' pages with over 10,000 followers. @logans\_pawsome\_friends is currently a low popularity page, but growing rapidly. In just over two weeks of the creation of the page, it has garnered 55 followers completely organically, without any promotion, either on Instagram or otherwise. 

%% ADD IMAGE: Images with highest IES
\begin{figure}[t]
  \centering
   \includegraphics[width=0.99\linewidth]{ 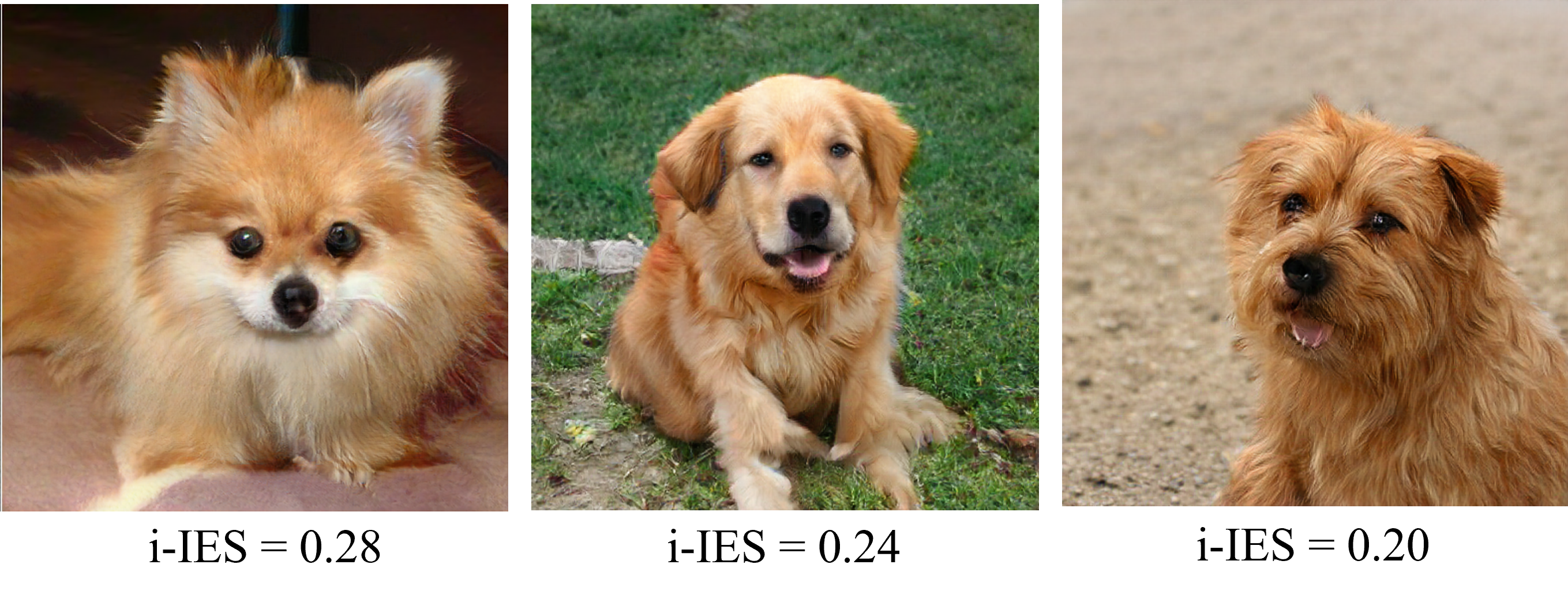}

   \caption{Images which have the highest engagement score (i-IES) so far.}
   \label{fig:ies}
\end{figure}

The user engagement of the page, as measured by the Instagram Engagement Score (p-IES) is 1.254 at the time of writing this paper.

\subsection{Posting Low-Fidelity Images}

While only the most realistic images were selected for posting on the Instagram page initially, I experimented with uploading a random image generated by the model directly to the page. This was potentially risky because lower-quality images may hurt engagement for the specific image as well as the page in general. It would, nonetheless, be useful to measure the impact of image fidelity on user engagement.

The first such image to be uploaded, however, was taken down by Instagram shortly after uploading for violating its terms of use based on user reports. Interestingly, the image was flagged as ``Bullying or Harassment'' rather than ``Spam,'' perhaps indicating that the user viewed the disfigured image of the dog as animal abuse rather than as a fake image. Regardless, due to the likelihood of images being reported and the risk of the page being banned altogether, the experiment was subsequently abandoned.

%% ADD Image: Image which got deleted

\section{Analysis}

% Analysis (8 points): Make a direct comparison to baselines/previous works and explain your results. Explore the tradeoffs between approaches with theoretical basis.

%% Compare to Jain et al. and Brock et al.
My models yielded results that were either comparable or better than related tasks in prior work. The results are summarized in Table \ref{table:modelcompare}.

\begin{table}
\begin{center}
\begin{tabular}{|l|l|}
\hline
Model & IS \\
\hline\hline
DCGAN - 350 epochs \cite{jain2020performance} (Dog Images) & 5.281 \\
BigGAN - 50 epochs \cite{jain2020performance} (Dog Images) & 8.173 \\
\textit{DCGAN (350 epochs) (Dog Images)} & 7.879\\
\textit{DCGAN (Baseline) (Dog Images)} & 10.399\\
\textit{DCGAN (Baseline) (Dog and Cat Images)} & 7.484\\
\textit{BigGAN (Pretrained) (Dog Images)} & 94.331\\
\textit{BigGAN (Pretrained) (Dog and Cat Images)} & 67.500\\
\hline
\end{tabular}
\end{center}
\caption{Summary of model performance in terms of inception score. My implementations which use the curated dataset described in Section \ref{sec:dataset} are italicized.}
\label{table:modelcompare}
\end{table}

A key difference between prior work and my implementation is the choice of the dataset and related preprocessing. This underlines the fact that the model performance is sensitive to the quality of the training data. An important limitation is that the training time varies between the prior work and present study, and not all results are strictly comparable (except the first item and the third item in the table). Another limitation of this comparison is that it does not account for differences in hardware used for training the models.

While no prior work used images of both dogs and cats, using the comparison between the performance of my models on different data, it is clear that the models performed better when only generating dog images. This may be because of two reasons: (a) the addition of a new class of data makes the task inherently more complex, and (b) cat images, in particular, may be more difficult to generate. While (a) is a well-understood problem, I speculate that (b) may have a role to play because of the quality of results generated by the model. 

%% Add image: Cats!
\begin{figure}[t]
  \centering
   \includegraphics[width=0.99\linewidth]{ 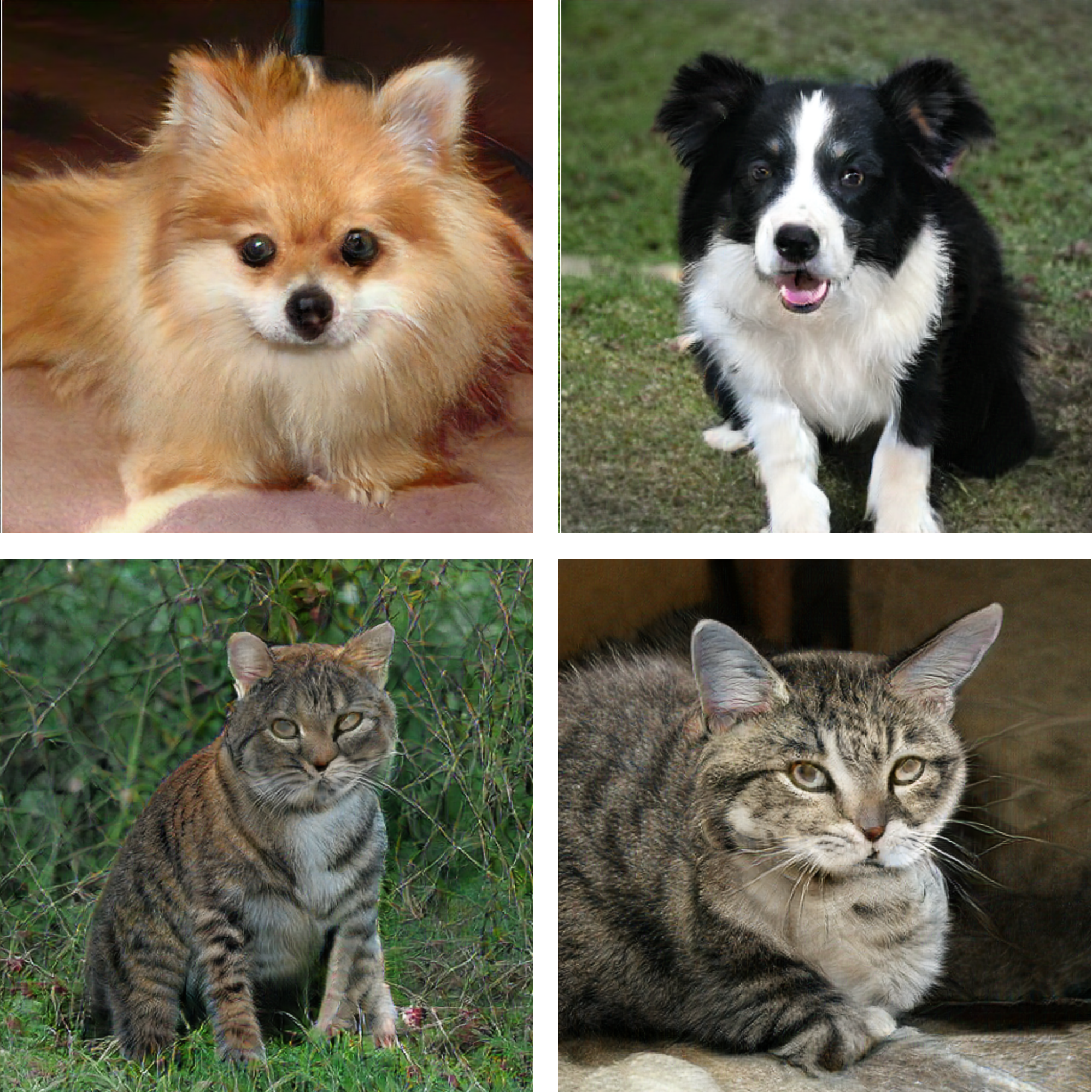}

   \caption{The model performs poorly in terms of generating eyes. Since the feature is more distinctive for cats than dogs, the top two images were considered fit for posting on Instagram, while the bottom two were not.}
   \label{fig:eyes}
\end{figure}

Specifically, in terms of interpretable features, the model tends to perform poorly in terms of generating realistic eyes for animals. This is true for both dog images and cat images. However, for dogs, the eyes are often uniformly colored and contain indistinct irises; as a result, these images may nevertheless seem indistinguishable from real images to humans. On the other hand, cats have distinctive irises. As a result, imperfect eyes can result in images not being suitable for human consumption altogether. Examples of this are shown in Figure \ref{fig:eyes}. This underlined a key failure of the model, as it was unable to generate high-quality cat images.

%% Compare user engagement across pages
The page-level engagement for @logans\_pawsome\_friends was compared against average (mean) page-level engagements for high-popularity, medium-popularity and low-popularity pages. For each category, $n = 20$ pages were sampled. The distribution of the number of followers roughly corresponded to the standard normal distribution centered around the midpoint of the range of followers for each category. The upper limit for the range in the case of ``high-popularity'' pages was set to 9.9M followers, which is the number of followers on the @jiffpom page, the most popular page with comparable content. The results are summarized in Table \ref{table:engagementcompare}.

\begin{table}
\begin{center}
\begin{tabular}{|l|l|}
\hline
Page / Page Type & p-IES \\
\hline\hline
High-popularity pages (average) & 0.025 \\
Medium-popularity pages (average) & 0.754 \\
Low-popularity pages (average) & 0.612\\
@logans\_pawsome\_friends & 1.254\\
\hline
\end{tabular}
\end{center}
\caption{Comparison of page-level engagement.}
\label{table:engagementcompare}
\end{table}

@logans\_pawsome\_friends drives more engagement than the average in each category. Additionally, it also has significantly higher engagement than the most popular page, @jiffpom, with p-IES $= 0.026$. This may be at least partly due to the fact that as the number of followers grows on a page, it is likely that a large number of followers are passive, with engagement driven only by a small subset of active users \cite{blaine2018we}. This may also explain why engagement drops for high-popularity pages in general. However, @logans\_pawsome\_friends, a low-popularity page, also drives more engagement compared to the average low-popularity page, which may be attributed to the fact that it is able to deliver content more consistently, which can drive engagement. 

The comparisons may suffer from sampling bias, especially in the low-popularity category, since these pages were sampled through Instagram-delivered results based on popular hashtags. Additionally, given the relatively small amount of images on @logans\_pawsome\_friends presently, it is not possible to isolate the effects of other variables that may drive engagement including, but not limited to, hashtags used with the images and time of the day of uploading the images. Regardless, the results show that GAN-generated images are capable of driving user engagement at levels that are at least comparable to those achieved by real images.

\section{Conclusion}

% Conclusion (3 points):  Summarize your results and what you have learned through your exploration. Address areas for future exploration and research.

In this study, I evaluated the quality of pet images generated by two different GAN architectures, DCGAN and BigGAN, and evaluated their applicability for use as therapeutic social media content. The model performance results were consistent with prior work presented in \cite{jain2020performance} in that the BigGAN model generated better-quality images. However, my experiments showed that the results may be further improved by the choice of the dataset. Additionally, using a pretrained model can have significant benefits, both in terms of training time and the quality of images generated. This contributed to the significant improvement in the BigGAN-generated images when compared to Jain et al.'s work \cite{jain2020performance}. 

The dog images generated by the models were consistently better than the cat images; as a result, only dog images were used for evaluating user engagement with these images. These images were uploaded to @logans\_pawsome\_friends, a newly-created Instagram account, and the engagement generated by the images was compared to that of similar pages' content. @logans\_pawsome\_friends achieved higher engagement than pages with real pet images on average, which underlines the applicability of the approach. This study, therefore, provides a basis for the creation of GAN-powered pet therapy pages, which can engage users at levels comparable to traditional pages, while allowing for large-scale content creation.

In the future, this study may be extended by exploring new techniques or augmenting the discussed techniques (by using new data or training for longer) to generalize the task to other pets (such as cats). Additionally, further data can be collected from the Instagram page over an extended period of time to understand the individual impact of different variables on driving user engagement and to reinforce confidence in the applicability of this approach as the page grows in popularity.

%-------------------------------------------------------------------------

{\small
\bibliographystyle{ieee_fullname}
\bibliography{ egpaper}
}

\end{document}